\documentclass{article} % For LaTeX2e
\usepackage{iclr2017_workshop,times}
\usepackage{hyperref}
\usepackage{url}

\usepackage[]{algorithm2e}
\usepackage{graphicx}
\usepackage{amsmath}
\usepackage{subcaption}
\usepackage{helvet}
\usepackage{courier}
\usepackage{amsthm}

\newtheorem{theorem}{Theorem}[section]

\begin{document}
% % The file aaai.sty is the style file for AAAI Press
% % proceedings, working notes, and technical reports.
% %
% \title{HiNet: Hierarchical Classification with Neural Network}
% % \author{AAAI Press\\
% % Association for the Advancement of Artificial Intelligence\\
% % 2275 East Bayshore Road, Suite 160\\
% % Palo Alto, California 94303 \\
% % }
% \author{Zhenzhou Wu \\ SAP Innovation Center, Singapore \\ ... \\ Address line
% \And
% Sean Saito \\ Yale-NUS College \\ Singapore, 138533 \\ \texttt{sean.saito@u.yale-nus.edu.sg}
% \And
% Daniel
% \And
% Sujoy
% \And
% ZhengXi}

\title{HiNet: Hierarchical Classification with Neural Network}

\author{Zhenzhou Wu \\ SAP Innovation Center, Singapore \\ Singapore 119958 \\ \texttt{zhenzhou.wu@sap.com}
\And
Sean Saito \\ Yale-NUS College \\ Singapore, 138533 \\ \texttt{sean.saito@u.yale-nus.edu.sg}}
% \And
% Xin Zheng \\ Nanyang Technological University, Singapore \\
% SAP Innovation Center, Singapore \\ Singapore 119958 \\ \texttt{xzheng008@e.ntu.edu.sg, xin.zheng@sap.com}
% \And
% Sujoy Roy \\ SAP Innovation Center, Singapore \\ Singapore 119958 \\ \texttt{sujoy.roy@sap.com}
% \And
% Daniel Dahlmeier \\ SAP Innovation Center, Singapore \\ Singapore 119958 \\ \texttt{d.dahlmeier@sap.com}
% }

% \author{Xin Zheng
% 	\\  Nanyang Technological University, Singapore
% 	\\  SAP Innovation Center, Singapore
% 	\\ \texttt{xzheng008@e.ntu.edu.sg, xin.zheng@sap.com}
% \And
% \author{Zhenzhou Wu \\ SAP Innovation Center, Singapore \\ Singapore 119958 \\ \texttt{zhenzhou.wu@sap.com}
% \And
% \author{Zhenzhou Wu \\ SAP Innovation Center, Singapore \\ Singapore 119958 \\ \texttt{zhenzhou.wu@sap.com}
% \And
% \author{Zhenzhou Wu \\ SAP Innovation Center, Singapore \\ Singapore 119958 \\ \texttt{zhenzhou.wu@sap.com}}
% \And
% Daniel
% \And
% Sujoy
% \And
% ZhengXi}

\maketitle
\begin{abstract}
Traditionally, classifying large hierarchical labels with more than 10000 distinct traces can only be achieved with flatten labels. Although flatten labels is feasible, it misses the hierarchical information in the labels. Hierarchical models like HSVM by \cite{vural2004hierarchical} becomes impossible to train because of the sheer number of SVMs in the whole architecture. We developed a hierarchical architecture based on neural networks that is simple to train. Also, we derived an inference algorithm that can efficiently infer the MAP (maximum a posteriori) trace guaranteed by our theorems. Furthermore, the complexity of the model is only $O(n^2)$ compared to $O(n^h)$ in a flatten model, where $h$ is the height of the hierarchy.
\end{abstract}

\section{Introduction}

Large hierarchical classification with more than 10000 categories is a challenging task (\cite{partalas2015lshtc}). Traditionally, hierarchical classification can be done by training a classifier on the flattened labels (\cite{babbar2013flat}) or by training a classifier at each hierarchical node (\cite{silla2011survey}), whereby each hierarchical node is a decision maker of which subsequent node to route to. However, the second method scales inefficiently with the number of categories ($> 10000$ categories). Models such as hierarchical-SVM (\cite{vural2004hierarchical}) becomes difficult to train when there are 10000 SVMs in the entire hierarchy. Therefore for large number of categories, the hierarchy tree is flattened to produce single labels. While training becomes easier, the data then loses prior information about the labels and their structural relationships.

In this paper, we model the large hierarchical labels with layers of neurons directly. Unlike the traditional structural modeling with classifier at each node, here we represent each label in the hierarchy simply as a neuron.

% \subsection{Transfer Learning}
% % \todo{zhenzhou}
% Transfer learning is a well studied field which typically involves the transfer of knowledge from 1. feature \cite{} 2 model parameters \cite{}. Figure \ref{} shows typical settings of different types of transfer learning. In this work, we propose a new model (Figure \ref{}) of transfer learning by using the cost function of different labels to transfer the knowledge to target label, assuming that other labels also contain some information of the input feature. Moreover, we look at labels with conditional dependencies, $p(y_1, y_2,\dots,y_k | X)$, and $p(y_k | y_{k-1}, y_{k+1})$ and explore how other labels $y_1, y_{k-1},\dots,y_{k+1},\dots$ transfer the knowledge through cost function to the target label $y_k$.

% Furthermore, we investigate the strength of the knowledge transfer with respect to the strength of the conditions between labels $p(y_k | y_{k-1}, y_{k+1})$ by turning off some of the labels from adjacent level, and weakening the connections between labels and see how this weakening affects knowledge transfer through combined cost function.

\section{Model}
HiNet has different procedures for training and inference. During training, as illustrated in Figure \ref{fig:training}, the model is forced to learn MAP (Maximum a Posteriori) hypothesis over predictions at different hierarchical levels independently. Since the hierarchical layers contain shared information as child node is conditioned on the parent node, we employ a combined cost function over errors across different levels. A combined cost allows travelling of information across levels which is equivalent to transfer learning between levels.

During inference, after predicting the posterior distribution at each level of the hierarchy, we employ a greedy downpour algorithm to efficiently infer the MAP (Maximum a Posteriori) hierarchical trace (Figure \ref{fig:inference}) from the posterior predictions at each level (Figure \ref{fig:training}).

% the connected together by a combined cost function for transfer learning. Since hierarchical levels contain shared information between levels, it makes sense to do transfer learning across different levels, which we model it with a combined cost function. Also we will described an greedy inference method for efficiently inferencing the hierarchical labels.

% \subsection{One Auxiliary Case}
% Considering the case with one auxiliary label, we can reduce Equation \ref{eqn:comberror} into
% \begin{equation}
%     E = \frac{1}{n}\sum_{i=1}^n\Big[\big(f(x_i) - t_{i}\big)^2 + \big(g(x_i) - a_{i}\big)^2\Big]
% \end{equation}

\subsection{Hierarchy as Layers of Neurons}

\begin{figure}
\hspace{2cm}
\begin{subfigure}{0.3\textwidth}
\includegraphics[width=4cm]{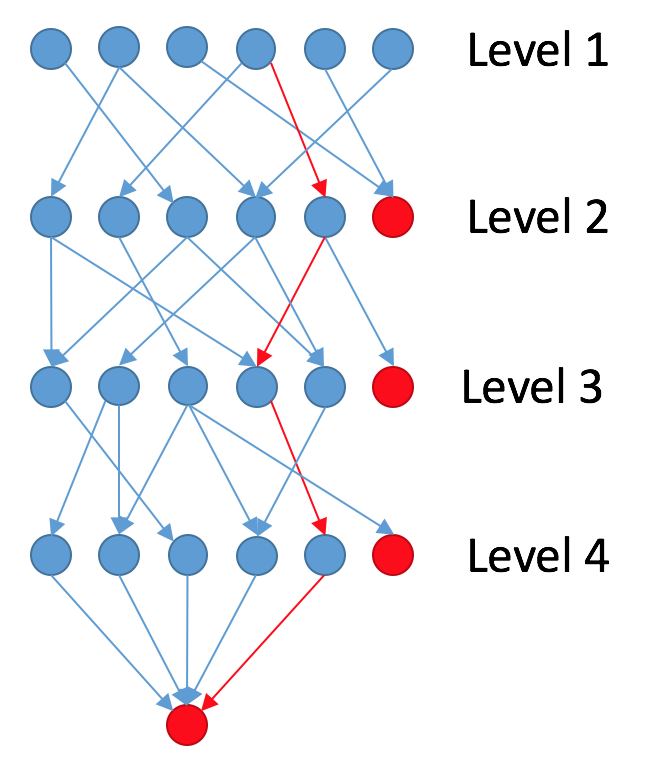}
\caption{DAG}
\label{fig:dag}
\end{subfigure}
\hspace{1cm}
\begin{subfigure}{0.3\textwidth}
\includegraphics[width=4cm]{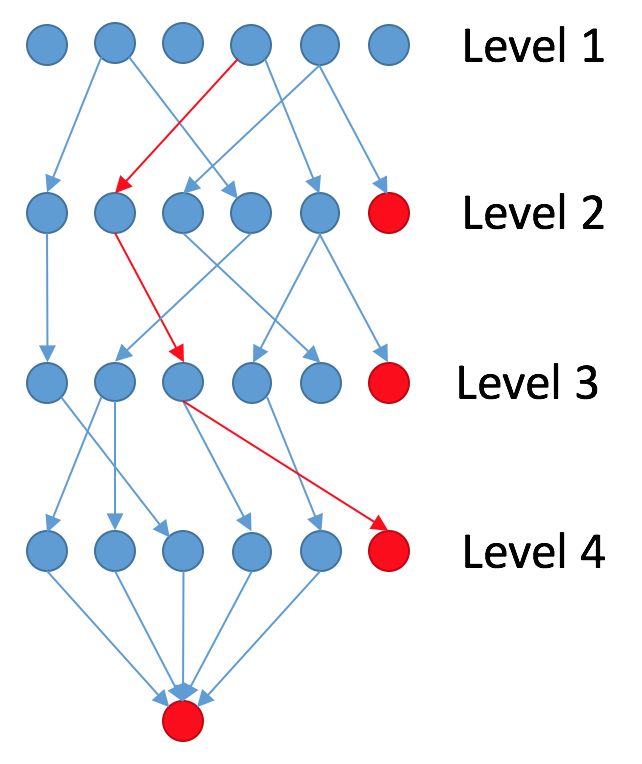}
\caption{Tree}
\label{fig:tree}
\end{subfigure}
\caption{Neural network for modeling hierarchical relationships. Figure \ref{fig:dag} shows a DAG (Directed Acyclic Graph) where a child neuron is possible to have more than one parents versus Figure \ref{fig:tree} showing a tree where each child neuron only belongs to one parent. The path will end in a stop neuron (red neuron).}
\label{fig:mask}
\vspace{-0.5cm}
\end{figure}

Using layers of neurons to model hierarchy is very efficient and flexible. It can be easily used to model a Directed Acyclic Graph (DAG) (Figure \ref{fig:dag}) or a Tree (Figure \ref{fig:tree}) by masking out the unnecessary connections. Unlike node-based architecture (\cite{silla2011survey,vens2008decision, dumais2000hierarchical}), whereby each node is an object with pointers to its child and parent, and takes up large memory, neural network models the connections as compact matrix which takes up much less memory. In order to model hierarchies of different length, we append a stop neuron (red neuron in Figure \ref{fig:mask}) at each layer. So a top-down path will end when it reaches the stop neuron.

% \subsection{Transfer Learning}

% \begin{figure}[h]
% \centering
%     \includegraphics[width=0.5\linewidth]{images/training.png}
%     \caption{Transfer learning with combined cost function}
%     \label{fig:training}
% \end{figure}

% For datasets with hierarchical structure, often there is strong correlations for labels across levels (Figure \ref{fig:mask}), therefore it is important to transfer this shared information between levels. We do it using a combined cost function.

\subsection{Training}

\begin{figure}[t]
% \vspace{-2cm}
\centering
% \begin{subfigure}{0.4\textwidth}
\centering
    \includegraphics[width=8cm]{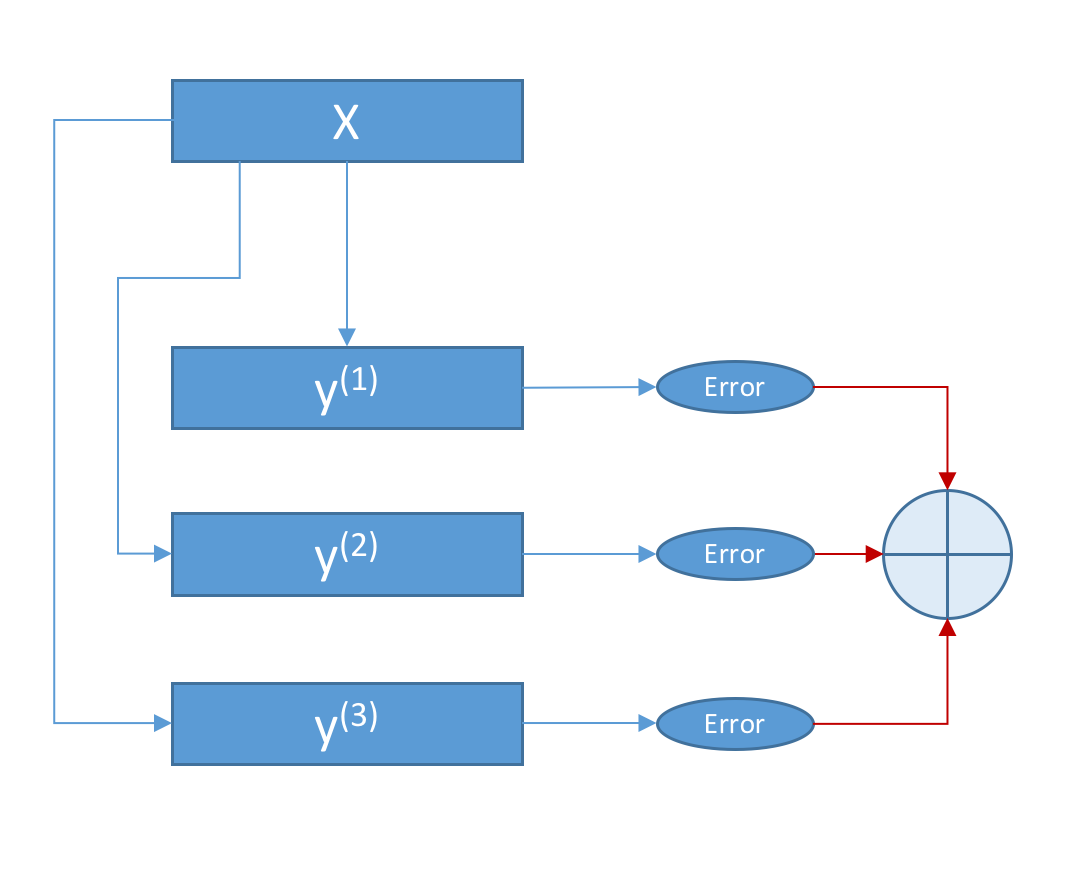}
    % \caption{Training: Transfer learning with combined cost function}
    \label{fig:training}
% \end{subfigure}
% \hspace{-2cm}
% \begin{subfigure}{0.5\textwidth}
% \centering
%     \includegraphics[width=4cm]{images/inference2.png}
%     \caption{Inference: Downpour Algorithm}
%     \label{fig:inference}
% \end{subfigure}
    \caption{The above figure illustrates the architecture during training.}
\label{fig:training}
\end{figure}

Figure \ref{fig:training} shows the model for transfer learning with combined cost function. Given an input feature $\mathbf{X}$ with multiple levels of outputs $\{\mathbf{y}^{(1)}, \mathbf{y}^{(2)}, \dots, \mathbf{y}^{(n)}\}$, where the outputs may have inter-level dependencies $p(\mathbf{y}^{(k)}|\mathbf{y}^{(1:k-1)}, \mathbf{y}^{(k+1:n)})$. For each output level from network $f_{\theta_k}(\mathbf{X}) = \mathbf{y}^{(k)}$ and its corresponding label $ \tilde{\mathbf{y}}^{(k)}$. The combined cost is defined as $E = \sum_{k}^n\Big(\tilde{\mathbf{y}}^{(k)} -f_{\theta_k}(\mathbf{X})\Big)^2$ allows the parameters $\theta_k$ from different levels to exchange knowledge.

% \begin{equation}
    % \label{eqn:cost}
    % E = \sum_{k}\Big(\tilde{\mathbf{y}}^{(k)} -\mathbf{y}^{(k)}\Big)^2
% \end{equation}

% If we zoom in at the $k$-th level, the error with respect to the $k$-th level is
% \begin{equation}
% \label{eqn:comberror}
% \begin{split}
%     E &= \Big(\tilde{\mathbf{y}}^{(k)} -\mathbf{y}^{(k)}\Big)^2 + \sum_{j\neq k}\Big(\tilde{\mathbf{y}}^{(j)} -\mathbf{y}^{(j)}\Big)^2 \\
%     &= \Big(\tilde{\mathbf{y}}^{(k)} -\mathbf{y}^{(k)}\Big)^2 + E(\theta_{1:k-1}, \theta_{k+1:n})
% \end{split}
% \end{equation}
% where the second term acts as a message passing from other levels.

% In the limit of large dataset with continuous space, we can express Equation \ref{eqn:cost} as

% \begin{equation}
%     \label{eqn:contcost}
%     E = \int p(\tilde{\mathbf{y}}^{(1:n)})\sum_{k=1}^n\Big(\tilde{\mathbf{y}}^{(k)} -\mathbf{y}^{(k)}\Big)^2 d\tilde{\mathbf{y}}^{(1)} d\tilde{\mathbf{y}}^{(2)} \dots d\tilde{\mathbf{y}}^{(n)}
% \end{equation}

% \subsection{Joint Entropy}
% To understand how transfer learning benefits from shared information, we can quantify the joint entropy of two output distributions between two levels $l$ and $m$
% \begin{equation}
%     H(Y_a^{(l)}, Y_b^{(m)}) = -\sum_{y_a^{(l)},y_b^{(m)}}p(y_a^{(l)},y_b^{(m)})\log p(y_a^{(l)},y_b^{(m)})
% \end{equation}
% whereby we want to understand how the joint entropy of target label and auxiliary label correlates with the knowledge transfer by the combined error.

\subsection{Inference}

\begin{figure}[h]
% \vspace{-2cm}
\centering
% \begin{subfigure}{0.4\textwidth}
% \centering
%     \includegraphics[width=5cm]{images/training.png}
%     \caption{Training: Transfer learning with combined cost function}
%     \label{fig:training}
% \end{subfigure}
% \hspace{-2cm}
% \begin{subfigure}{0.5\textwidth}
\centering
    \includegraphics[width=7cm]{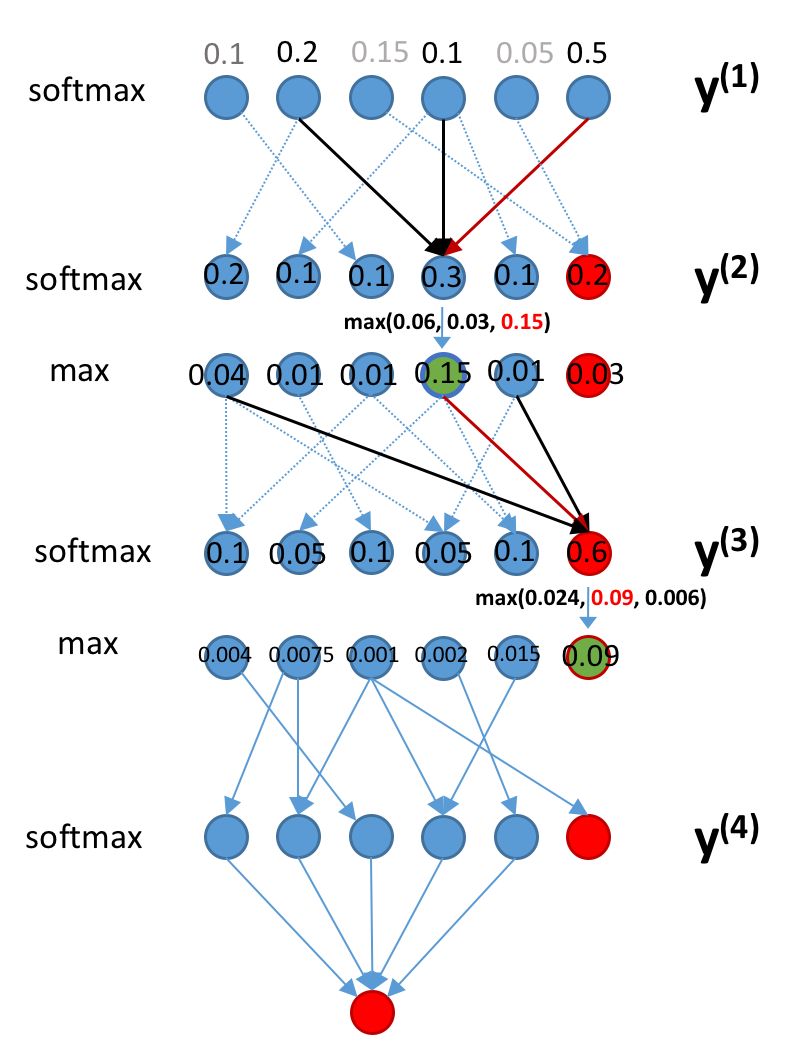}
    \caption{Inference: Downpour Algorithm}
    % \label{fig:inference}
% \end{subfigure}
    \caption{The above figure illustrates the downpour algorithm (Algorithm \ref{algo:inference}) for deriving the MAP trace.}
\label{fig:inference}
\end{figure}

% \begin{lemma}
% Given two line segments whose lengths are $a$ and $b$ respectively there is a
% real number $r$ such that $b=ra$.
% \end{lemma}

% The combined probability of predictions $p(y^{(1)}, y^{(2)}, \dots, y^{(k)})$ at levels $1$ to $k$ follows Markov chain whereby the prediction at level $k$ is only conditioned on its parent level $k-1$ defined as $p(y^{(k)} | y^{(k-1)})$, the full Markov expansion of the combined probability of predictions is defined as

% \begin{equation}
%     \label{eqn:conditional}
%     p(y^{(1)}, y^{(2)}, \dots, y^{(k)}) = p(y^{(1)}) \prod_{i=2}^k p(y^{(i)} | y^{(i-1)})
% \end{equation}

% To find the MAP of the combined predictions of all levels, we can do it greedily by finding the MAP of the predictions for all the previous layers multiply by the maximum of the conditional likelihood at this layer as described by Equation \ref{eqn:greedy}. After computing the MAP greedily, we can back-trace the predictions as described by Algorithm \ref{algo:inference} and illustrated in Figure \ref{fig:inference}.

% \begin{equation}
% \label{eqn:greedy}
% \begin{split}
%     &\max{p(y^{(1)}, y^{(2)}, \dots, y^{(k)})} \\
%     =&\max \big[p(y^{(k)}|y^{(k-1)}) \times p(y^{(1)}, y^{(2)}, \dots, y^{(k-1)})\big] \\
%     =&\max p(y^{(k)}|y^{(k-1)}) \times \max{p(y^{(1)}, y^{(2)}, \dots, y^{(k-1)})}
% \end{split}
% \end{equation}

% The inference algorithm resembles that of the inference in HMM \cite{} with a forward pass that computes the MAP up to a stop neuron or the last layer and a backward pass that outputs the trace of predictions for the MAP.

\paragraph{Downpour Algorithm}
% \begin{figure}[t]
% % \vspace{-2cm}
% \centering
% \begin{subfigure}{0.4\textwidth}
% \centering
%     \includegraphics[width=5cm]{images/training.png}
%     \caption{Training: Transfer learning with combined cost function}
%     \label{fig:training}
% \end{subfigure}
% % \hspace{-2cm}
% \begin{subfigure}{0.5\textwidth}
% \centering
%     \includegraphics[width=4cm]{images/inference2.png}
%     \caption{Inference: Downpour Algorithm}
%     \label{fig:inference}
% \end{subfigure}
%     \caption{Figure \ref{fig:training} illustrates the architecture during training. Figure \ref{fig:inference} illustrates the downpour algorithm (Algorithm \ref{algo:inference}) for deriving the MAP trace.}
% \label{fig:hinet}
% \end{figure}

% \hrule
\begin{algorithm}[h]
% \hrule
% \vspace{-0.4cm}

% \hrule

Given output posteriors $\mathbf{y}^{(l)} = \{y_a^{(l)}\}$ at each layer $l \in \{1, \dots, K\}$

% Given input $\mathbf{x}$ and output $\mathbf{y}^{(i)} = f_{i}(\mathbf{x})$ for $i=1\dots k$
Define $T^{(l)}_i = [\mbox{ }]$ as MAP trace of neuron $i$ at level $l$

% $child\_parent = \{\}$

$T^{(1)}_i = [i]$

\For{\mbox{layer} $l=2$ to $k$}{

    Find MAP parent $A = \arg\max_a y_b^{(l)} y_a^{(l-1)}$

    $T_b^{(l)} = T_A^{(l-1)}.append(b)$

    Update $y_b^{(l)}$  = $\max_a y_b^{(l)} y_a^{(l-1)}$

    % Find max child state $B = \arg\max_b y_b^{(l)}$

    % $child\_parent[B] = A$

    % \If{$B$ is stop neuron}
    % {
        % \For{layer m=l-1 to 2}
        % {
        %
        % }

        % Return the MAP trace

        % $trace = [B]$

        % \Repeat{B has no parent}{

        %     Find parent $B = child\_parent[B]$

        %     $trace.append(B)$}

        % % \Until

        % \Return  $trace$
    % }
}

Find MAP level for stop neuron $s$

$L = \arg\max_l y_s^{(l)}$

\Return $T_s^{(L)}$

\vspace{0.2cm}
\hrule
\vspace{0.3cm}
\caption{Downpour algorithm for inferencing the MAP trace.}
\label{algo:inference}
\end{algorithm}

During inference, as illustrated in Figure \ref{fig:inference}, the model will output a normalized probability distribution at each level $\{\mathbf{y}^{(1)}, \mathbf{y}^{(2)}, \dots, \mathbf{y}^{(k)}\}$, where $\mathbf{y}^{(k)} = \{y_{a_{k}}^{(k)}\}$ is a vector with indexes $a_k \in \{1,\dots, n\}$, where $n$ is the size of layer $k$, and $\sum_{a_k}y_{a_k}^{(k)}=1$. From the second level onwards, we include a stop neuron (red neuron) which is used for stopping the hierarchical trace. The path of the trace from top down ends in a stop neuron (red neuron). Define the MAP trace up to level $k$ which ends at stop neuron $s$ as $T^{(k)}_{a_k=s} = \arg\max\limits_{a_{1:k-1}} p(a_1, a_2, \dots, a_{k-1}, a_k=s)$. The objective of the downpour in finding the MAP from the hierarchy is equivalent to finding the maximum MAP trace out of all MAP traces that ends in a stop neuron from different levels which is $T^{(L)}_{a_L=s}$ where $L = \arg\max_k T^{(k)}_{a_k=s}$.
The probability of MAP trace at level $k$ can be derived greedily from the MAP trace at level $k-1$ as
\begin{equation}
p(T_{a_k}^{(k)})=\max_{a_{k-1}}p(a_k|a_{k-1})p(T_{a_{k-1}}^{(k-1)})
\label{eqn:greedy}
\end{equation}

From Equation \ref{eqn:greedy}, we can derive Theorems \ref{thm:general}-\ref{thm:downpour} which prove that Downpour Algorithm will always yield the MAP trace of $T^{(L)}_{a_L=s} = \max_n p(T_{a_n=s}^{(n)}) \geq p(a_1, a_2,\dots,a_m=s)$ $\forall m$.

% The objective is to find MAP trace in the hierarchy which is the maximum of

% Which is typically the MAP trace that ends in a stop neuron.

% To do so, we downpour probabilities greedily down each level and update the combined probabilities,

% $p_{l}(a_1, a_2, \dots, a_l) = p(a_l | a_{l-1})\max_{a_{l-1}}p_{MAP}(a_1, a_2, \dots, a_{l-1})$.

% For level $k$, to compute $\max p(y_{a_k}^{(k)}, y_{a_{k-1}}^{(k-1)}, \dots, y_{a_1}^{(1)}) = \max \limits_{a_{k-1}} \big[p(y_{a_k}^{(k)}) p(y_{a_{k-1}}^{(k-1)}, \dots, y_{a_1}^{(1)})\big]$ where $a_{k-1}$ is a child of $a_{k}$. After updating the maximum combined probabilities for each neuron position $a_k$ in level $k$, we find the position $a_k$ that gives the maximum combined probabilities (green neuron), if $a_k$ is a stop neuron, then we stop down-pouring and back-trace the path (red arrows) that gives the best combined probabilities.

% During inference, we want to find the MAP of the combined predictions across all levels $p(y_{a_1}^{(1)}, y_{a_2}^{(2)}, \dots, y_{a_k}^{(k)})$ where $y_k = f_k(x)$ is the predicted output at level $k$. Since the next level is conditioned on the previous level, therefore we can express the combined likelihood of all levels as conditional likelihood at each level.

% \begin{lemma}(Shorter Theorem)
% \label{lem:shorter}
% For a greedy downpour that ends at a stop neuron at level $n$ with sequence $S_n = \{y^{(1)},y^{(2)},\dots,y^{(n)}\}$, for every sequence $S_m = \{y^{(1)},y^{(2)},\dots,y^{(m)}\}$ where $m < n$, then $p(S_m) < p(S_n)$.
% \begin{proof}
% refer to appendix
% \end{proof}
% \end{lemma}

\begin{theorem}
\label{thm:general}
For a greedy downpour that ends at a stop neuron at level $n$ with MAP trace $T_{a_n}^{(n)}$, then $p(S_m) \leq p(T_{a_n}^{(n)})$ for every sequence $S_m=\{a_{1}, a_{2}, \dots, a_n, \dots, a_m\}$ of $m \geq n$ that pass through $a_n$ or ends at $a_n$. \textit{Refer to Appendix A for proof.}

% $S_n = \{y^{(1)},y^{(2)},\dots,y^{(n)}\}$, for every sequence $S_m = \{y^{(1)},y^{(2)},\dots,y^{(m)}\}$ where $m > n$, then $p(S_m) < p(S_n)$.
% \begin{proof}

% \end{proof}
\end{theorem}

\begin{theorem}
\label{thm:longer}
For a MAP trace $T_{a_n=s}^{(n)}$ that ends at stop neuron $s$ at level $n$ such that $p(T_{a_n=s}^{(n)}) \geq p(T_{a_n}^{(n)})$ $\forall a_n \neq s$, then $p(S_m) \leq p(T_{a_n=s}^{(n)})$ for every sequence $S_m=\{a_{1}, a_{2}, \dots, a_m\}$ of $m>n$. \textit{Refer to Appendix A for proof.}
% \begin{proof}
% refer to appendix
% \end{proof}
\end{theorem}

\begin{theorem}
\label{thm:downpour}
The maximum of the MAP traces that end in a stop neuron from each level is $T_{a_k=s}^{(L)}$ where $L=\arg\max_k p(T_{a_k=s}^{(k)})$ is the MAP for the hierarchy, which means $p(T_{a_k=s}^{(L)}) \geq p(S_m)$ $\forall m$. \textit{Refer to Appendix A for proof.}
% \begin{proof}
% refer to appendix
% \end{proof}
\end{theorem}

% \begin{theorem}(Downpour Optimality Theorem)
% \label{thm:downpour}

% \begin{proof}
% refer to appendix
% \end{proof}
% \end{theorem}

% \begin{algorithm}[t]
% \hrule
% \vspace{0.2cm}
% Given input $\mathbf{x}$ and output $\mathbf{y}^{(i)} = f_{i}(\mathbf{x})$ for $i=1\dots k$

% hier = []

% \For{$i=1$ to $k-1$}{
%     $L_{ab} \gets y^{(i)}_a \otimes y^{(i+1)}_{b}$

%     $L_{ab} \gets L_{ab} \times M^{(i)}_{ab}$ \quad \# $M^{(i)}_{ab}$ is masking

%     $y^{(i+1)}_b = \max\limits_{a}{L_{ab}}$

%     $m_b^{(i+1)} = \arg\max\limits_{a}{L_{ab}}$

%     hier.append($m_b^{(i+1)}$)

%     $g = \arg\max\limits_b y^{(i+1)}_{b}$

%     \If{$g$ is stop neuron}
%     {
%         trace = []

%         \For{$\mathbf{m}$ in reverse(hier)}{
%             $g = m_g$

%             trace.append($g$)
%         }
%         \Return trace
%     }
% }

% trace = [$g$]

% \For{$\mathbf{m}$ in reverse(hier)}{
%     $g = m_g$

%     trace.append($g$)
% }
% \Return trace
% \vspace{0.2cm}
% \hrule
% \vspace{0.3cm}
% \caption{Downpour algorithm for inferencing the MAP of predictions across all levels.}
% \label{algo:inference}
% \end{algorithm}

% \subsection{Masking}
% During

\section{Results and Conclusion}

% \subsection{BioASQ}

% \subsection{}

% \subsection{CCTL}

% \begin{table}[h]
% \begin{center}
% \begin{tabular}{|c|c|c|}
% \hline
%                      & Level 1 & Level 2 &
% \hline
% with transfer learning    &d         &   d      &
% \hline
% without transfer learning & d        &  d       &
% \hline
% \end{tabular}
% \end{center}
% \end{table}

% \subsection{}

\begin{table}[h]
\begin{center}
\begin{tabular}{|c|c|c|}
\hline
                     & DMOZ (Tree)   & No. of Params \\ \hline
Flatten Network      & 39.2         & $O(kn^h)$ \\ \hline
HiNet               &  41.4         & $O(kn + hn^2)$ \\ \hline
% \caption{Results on DMOZ dataset \cite{}}
\end{tabular}
\end{center}
\caption{Accuracy on the DMOZ dataset (\cite{partalas2015lshtc}) with 11947 classes. $k$: dimension of the first feature layer connected to the first hierarchical layer. $n$: dimension of each hierarchical layer. $h$: height of the hierarchy. For a fully dense hierarchy, the total number of classes is $n^h$.
}
\end{table}
We compared HiNet with a Flatten Network which have the same architecture except the output layer for HiNet is hierarchical as illustrated in Figure \ref{fig:training} and flatten for Flatten Network. The number of outputs for Flatten Network corresponds to the number of classes in the dataset. From the results, HiNet out-performs Flatten Network for both a Tree hierarchical dataset with much lesser parameters. We see that the number of parameters in the classification layer for Flatten Network is exponential to the maximum length of the trace. Thus for very deep hierarchies, the number of parameters in Flatten Network will be exponentially large while HiNet is always polynomial. This makes HiNet not only better architecture in terms of accuracy but also way more efficient in parameters space.

% \section{Conclusion}
% \nocite{*}

\newpage
\bibliography{iclr2017_workshop}

\begin{thebibliography}{6}
\providecommand{\natexlab}[1]{#1}
\providecommand{\url}[1]{\texttt{#1}}
\expandafter\ifx\csname urlstyle\endcsname\relax
  \providecommand{\doi}[1]{doi: #1}\else
  \providecommand{\doi}{doi: \begingroup \urlstyle{rm}\Url}\fi

\bibitem[Babbar et~al.(2013)Babbar, Partalas, Gaussier, and
  Amini]{babbar2013flat}
Rohit Babbar, Ioannis Partalas, Eric Gaussier, and Massih-Reza Amini.
\newblock On flat versus hierarchical classification in large-scale taxonomies.
\newblock In \emph{Advances in Neural Information Processing Systems}, pp.\
  1824--1832, 2013.

\bibitem[Dumais \& Chen(2000)Dumais and Chen]{dumais2000hierarchical}
Susan Dumais and Hao Chen.
\newblock Hierarchical classification of web content.
\newblock In \emph{Proceedings of the 23rd annual international ACM SIGIR
  conference on Research and development in information retrieval}, pp.\
  256--263. ACM, 2000.

\bibitem[Partalas et~al.(2015)Partalas, Kosmopoulos, Baskiotis, Artieres,
  Paliouras, Gaussier, Androutsopoulos, Amini, and Galinari]{partalas2015lshtc}
Ioannis Partalas, Aris Kosmopoulos, Nicolas Baskiotis, Thierry Artieres, George
  Paliouras, Eric Gaussier, Ion Androutsopoulos, Massih-Reza Amini, and Patrick
  Galinari.
\newblock Lshtc: A benchmark for large-scale text classification.
\newblock \emph{arXiv preprint arXiv:1503.08581}, 2015.

\bibitem[Silla~Jr \& Freitas(2011)Silla~Jr and Freitas]{silla2011survey}
Carlos~N Silla~Jr and Alex~A Freitas.
\newblock A survey of hierarchical classification across different application
  domains.
\newblock \emph{Data Mining and Knowledge Discovery}, 22\penalty0
  (1-2):\penalty0 31--72, 2011.

\bibitem[Vens et~al.(2008)Vens, Struyf, Schietgat, D{\v{z}}eroski, and
  Blockeel]{vens2008decision}
Celine Vens, Jan Struyf, Leander Schietgat, Sa{\v{s}}o D{\v{z}}eroski, and
  Hendrik Blockeel.
\newblock Decision trees for hierarchical multi-label classification.
\newblock \emph{Machine Learning}, 73\penalty0 (2):\penalty0 185, 2008.

\bibitem[Vural \& Dy(2004)Vural and Dy]{vural2004hierarchical}
Volkan Vural and Jennifer~G Dy.
\newblock A hierarchical method for multi-class support vector machines.
\newblock In \emph{ICML}, pp.\  105. ACM, 2004.

\end{thebibliography}
\bibliographystyle{iclr2017_workshop}

% \appendix
\newpage
\section{Appendix A}
\paragraph{Proof for Theorem \ref{thm:general}}

For a greedy downpour that ends at a stop neuron at level $n$ with MAP trace $T_{a_n}^{(n)}$, then $p(S_m) \leq p(T_{a_n}^{(n)})$ for every sequence $S_m=\{a_{1}, a_{2}, \dots, a_n, \dots, a_m\}$ of $m \geq n$ that pass through $a_n$ or ends at $a_n$.

\begin{proof}
% \begin{equation}
for $m=n$, that is $p(T_{a_n}^{(n)}) \geq p(S_n)$. By definition $T_{a_{1}}^{(1)} = a_1$
\begin{align}
\begin{split}
p(T_{a_n}^{(n)}) &= \max_{a_{n-1}}p(a_n | a_{n-1})p(T_{a_{n-1}}^{(n-1)}) \\
&= \max_{a_{n-1}}p(a_n | a_{n-1})  \max_{a_{n-2}}p(a_{n-1} | a_{n-2})  p(T_{a_{n-2}}^{(n-2)}) \\
&= \max_{a_{1:n-1}} p(a_n | a_{n-1}) p(a_{n-1} | a_{n-2}) \dots p(a_1) \\
&= \max_{a_{1:n-1}} p(a_n, a_{n-1}, \dots, a_1) \\
&\geq p(a_n, a_{n-1}, \dots, a_1)
\end{split}
\end{align}
% \end{equation}
for $m>n$, we just need to prove that $p(S_m) \leq p(S_n)$

\begin{align}
\begin{split}
p(S_m) &= p(S_n) p(a_{n+1}, a_{n+2}, \dots, a_m | S_n) \\
       &\geq p(S_n)
\end{split}
\end{align}
since $p(a_{n+1}, a_{n+2}, \dots, a_m | S_n) \leq 1$
\end{proof}

\paragraph{Proof for Theorem \ref{thm:longer}}

For a MAP trace $T_{a_n=s}^{(n)}$ that ends at stop neuron $s$ at level $n$ such that $p(T_{a_n=s}^{(n)}) \geq p(T_{a_n}^{(n)})$ $\forall a_n \neq s$, then $p(S_m) \leq p(T_{a_n=s}^{(n)})$ for every sequence $S_m=\{a_{1}, a_{2}, \dots, a_m\}$ of $m>n$.
\begin{proof}
from Theorem \ref{thm:general}, $p(S_m) \leq p(T_{a_n}^{(n)}) \leq p(T_{a_n=s}^{(n)})$.
\end{proof}

% \begin{proof}
% refer to appendix
% \end{proof}
% \end{theorem}

\paragraph{Proof for Theorem \ref{thm:downpour}}

The maximum of the MAP traces that end in a stop neuron from each level is $T_{a_k=s}^{(L)}$ where $L=\arg\max_k p(T_{a_k=s}^{(k)})$ is the MAP for the hierarchy, which means $p(T_{a_k=s}^{(L)}) \geq p(S_m)$ $\forall m$.

\begin{proof}
From Theorem \ref{thm:longer}, we have $p(T_{a_n=s}^{(n)}) \geq p(S_m)$ for every $m>n$, and from Theorem \ref{thm:general}, we have $p(T_{a_n=s}^{(n)}) \geq p(a_1, a_2, \dots, a_n=s)$. Therefore $\max_n p(T_{a_n=s}^{(n)}) \geq p(a_1, a_2,\dots,a_m=s)$ $\forall m$.
\end{proof}

% \begin{proof}
% refer to appendix
% \end{proof}
% \end{theorem}

% \begin{algorithm}[H]
% \hrule
% \vspace{0.2cm}

% Given output posteriors $\mathbf{y}^{(l)} = \{y_a^{(l)}\}$ at each layer $l \in \{1, \dots, K\}$

% % Given input $\mathbf{x}$ and output $\mathbf{y}^{(i)} = f_{i}(\mathbf{x})$ for $i=1\dots k$
% Define $T^{(l)}_i = [\mbox{ }]$ as MAP trace of neuron $i$ at level $l$

% % $child\_parent = \{\}$

% $T^{(1)}_i = [i]$

% \For{\mbox{layer} $l=2$ to $k$}{

%     Find MAP parent $A = \arg\max_a y_b^{(l)} y_a^{(l-1)}$

%     $T_b^{(l)} = T_A^{(l-1)}.append(b)$

%     Update $y_b^{(l)}$  = $\max_a y_b^{(l)} y_a^{(l-1)}$

%     % Find max child state $B = \arg\max_b y_b^{(l)}$

%     % $child\_parent[B] = A$

%     % \If{$B$ is stop neuron}
%     % {
%         % \For{layer m=l-1 to 2}
%         % {
%         %
%         % }

%         % Return the MAP trace

%         % $trace = [B]$

%         % \Repeat{B has no parent}{

%         %     Find parent $B = child\_parent[B]$

%         %     $trace.append(B)$}

%         % % \Until

%         % \Return  $trace$
%     % }
% }

% Find MAP level for stop neuron $s$

% $L = \arg\max_l y_s^{(l)}$

% \Return $T_s^{(L)}$

% \vspace{0.2cm}
% \hrule
% \vspace{0.3cm}
% \caption{Downpour algorithm for inferencing the MAP of predictions across all levels.}
% \label{algo:inference}
% \end{algorithm}

% \newpage
% \section{Appendix B}

% \begin{proof}
% Lemma \ref{lem:shorter}:
% \newline
% Assume there exists a shorter sequence $S_{n-1}$ such that $P(S_{n-1}) > P(S_{n})$.
% \end{proof}

\end{document}